\documentclass[letterpaper, 10 pt, conference]{ieeeconf}  
\IEEEoverridecommandlockouts                              
\usepackage{graphicx} 
\usepackage{subfigure}	
\usepackage[table]{xcolor}
\usepackage{diagbox}
\usepackage{multirow}
\usepackage{amsmath}
\usepackage{amsfonts,amssymb}
\usepackage{color}
\usepackage{cite}
\usepackage{booktabs}
\usepackage{pifont}
\usepackage[capitalize]{cleveref}
\usepackage{subcaption}
\usepackage{booktabs}

\title{\LARGE \bf
SDP: Spiking Diffusion Policy for Robotic Manipulation with\\ Learnable Channel-Wise Membrane Thresholds
}

\author{Zhixing Hou$^{1*}$, Maoxu Gao$^{1*}$, Hang Yu$^{1*}$, Mengyu Yang$^{1}$, Chio-In IEONG$^{1\dag}$ 
\thanks{$^{*}$ Equal contribution}
\thanks{$^{\dag}$ Corresponding author.}
\thanks{$^{1}$ Guangdong Institute of Intelligence Science and Technology, Hengqin, Zhuhai 519031, Guangdong, China. 
{\tt\small E-mail: houzhixing@gdiist.cn, yangchaoran@gdiist.cn}} 
}

\begin{document}
    \maketitle
    \begin{abstract}

    This paper introduces a Spiking Diffusion Policy (SDP) learning method for robotic manipulation by integrating Spiking Neurons and Learnable Channel-wise Membrane Thresholds (LCMT) into the diffusion policy model, thereby enhancing computational efficiency and achieving high performance in evaluated tasks. Specifically, the proposed SDP model employs the U-Net architecture as the backbone for diffusion learning within the Spiking Neural Network (SNN). It strategically places residual connections between the spike convolution operations and the Leaky Integrate-and-Fire (LIF) nodes, thereby preventing disruptions to the spiking states. Additionally, we introduce a temporal encoding block and a temporal decoding block to transform static and dynamic data with timestep $T_S$ into each other, enabling the transmission of data within the SNN in spike format. Furthermore, we propose LCMT to enable the adaptive acquisition of membrane potential thresholds, thereby matching the conditions of varying membrane potentials and firing rates across channels and avoiding the cumbersome process of manually setting and tuning hyperparameters. Evaluating the SDP model on seven distinct tasks with SNN timestep $T_S=4$, we achieve results comparable to those of the ANN counterparts, along with faster convergence speeds than the baseline SNN method. This improvement is accompanied by a reduction of 94.3\% in dynamic energy consumption estimated on 45nm hardware. 

    \end{abstract}
    
    \section{Introduction}
\label{sec:intro}

    The swift progress in embodied AI not only has sparked the revolutionary transformation in autonomous driving but also unveiled grand opportunities for applying such technologies in other forms of robotic systems, such as the next-generation intelligent robotic arms designed for versatile open-world manipulations. These intelligent arms will become increasingly integral in a wide array of sectors, such as industrial automation and healthcare, and are expected to expand the application scope e.g. into everyday life. The growing scope and complexity of applications have elevated the expectations and demands placed upon robotic arms. There is an increasing demand for capabilities such as robust open-world perception, common-sense reasoning and decision-making, swift adaptation to dynamic environments, efficient and precise trajectory planning, and last-but-not-least efficient computation in edge devices. 
    
    To address these challenges, large efforts have been paid to incorporating large language model and vision-language model~\cite{huang2023voxposer} into the system loop, developing efficient robotic manipulation models~\cite{liu2024robomamba} and conducting various model compression methods, such as pruning~\cite{han2015learning,fang2023depgraph}, knowledge distillation~\cite{sun2019patient}, and model quantization~\cite{li2023q,yuan2023rptq,hou2024fbpt}. Despite these advancements, a significant challenge remains in achieving efficient computation that maintains high accuracy and enables rapid response and low energy consumption, which is essential for the practical deployment of robotic arms in real-world scenarios.

    \begin{figure}[t]
		\centering
		\subfigure[]{
			\begin{minipage}[t]{0.44\linewidth}
				\centering
				\includegraphics[width=1\linewidth]{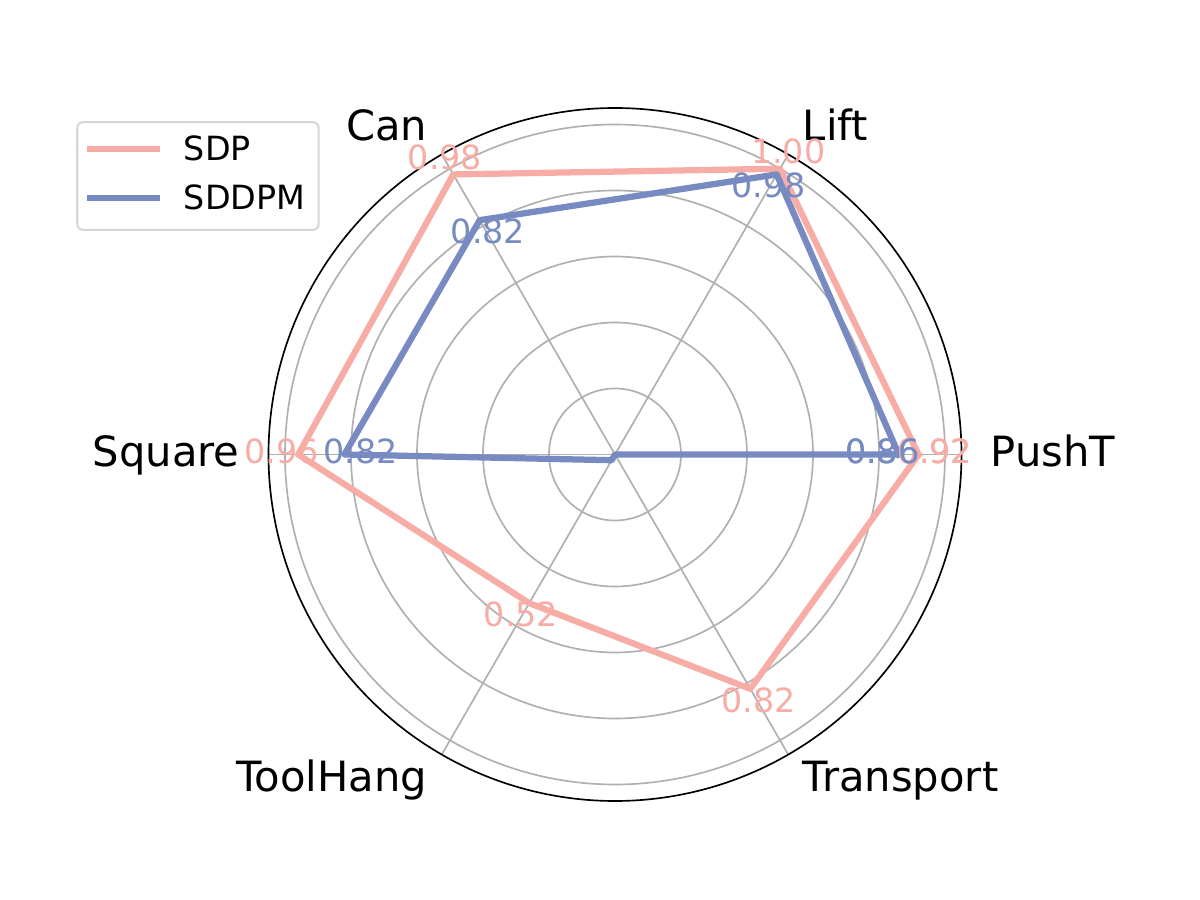}
                \label{fig:radar}
			\end{minipage}%
		}%
		\subfigure[]{
			\begin{minipage}[t]{0.5\linewidth}
				\centering
				\includegraphics[width=1\linewidth]{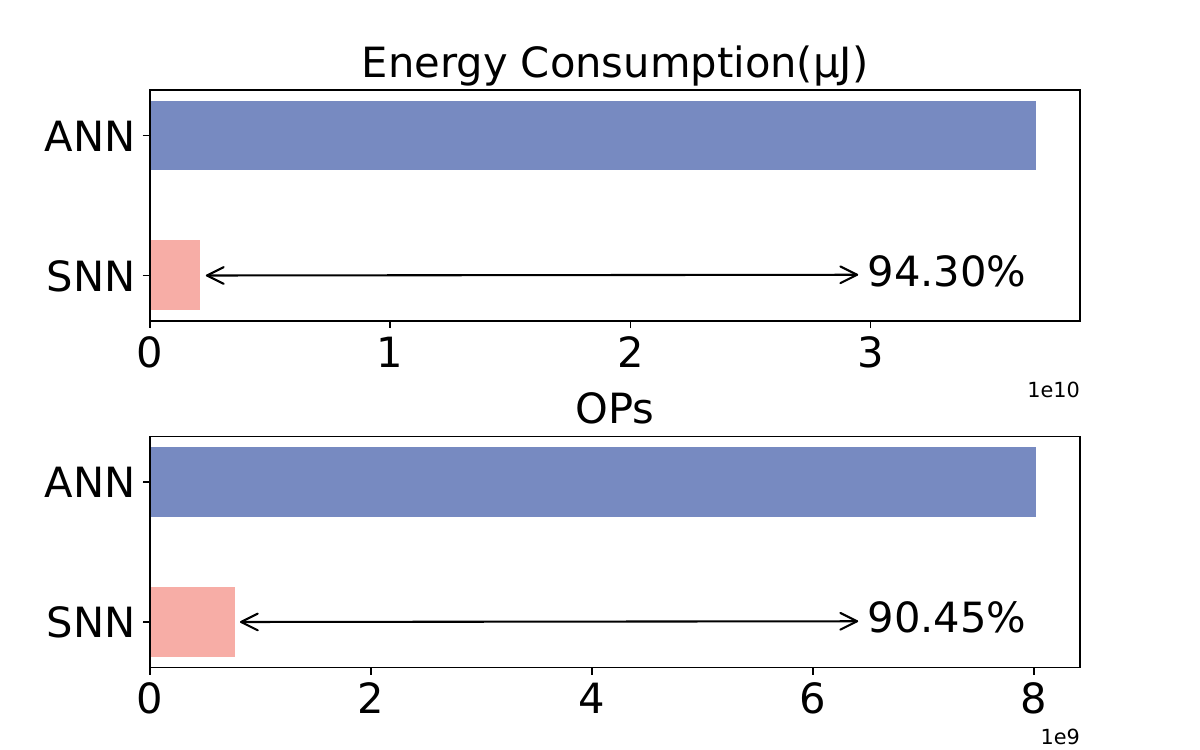}
                \label{fig:energy_ops}
			\end{minipage}%
		}
		\centering
		\caption{(a) Evaluation results on benchmarks. (b) Energy consumption and number of operations.}
		\label{fig:cover}
	\end{figure}

    Drawing inspiration from the extraordinary ability of biological brains to process large volumes of information efficiently, the field of neuromorphic computing~\cite{schuman2017survey,yu2020overview} using Spiking Neural Networks (SNNs)~\cite{maass1997networks} becomes an exciting, innovative, less-explored path to explore towards improving efficiency of neural network models. Designed to closely mimic the neural mechanisms found in nature, SNNs provide benefits like lower power consumption due to its binary spike representation and event-driven operation mechanism. SNN has been successfully applied in a variety of domains, encompassing tasks such as digit recognition~\cite{diehl2015unsupervised}, real-time sound source localization~\cite{liu2010biologically}, image classification~\cite{fang2021deep}, object detection~\cite{su2023deep}, enhancement of large language models (LLMs)~\cite{xing2024spikelm}, image generation~\cite{cao2024spiking}, and robotics~\cite{liu2021spiking}.

    In this paper, we introduce a Spiking Neural Network methodology for robotic manipulation, achieved by incorporating neuromorphic spiking neurons into the diffusion policy algorithm~\cite{chi2023diffusion}. This novel approach, designated as \textbf{S}piking \textbf{D}iffusion \textbf{P}olicy (\textbf{SDP}), significantly enhances computational efficiency while attaining performance on par with traditional Artificial Neural Networks.
    Specifically, the SDP utilizes a U-Net architecture as its backbone for noise prediction in the reverse diffusion process and incorporates Leaky Integrate-and-Fire (LIF) nodes as activation functions throughout the U-Net structure. Notably, we strategically place residual connections between the spike convolution operations and the LIF nodes, preventing disrupting the spiking states. Furthermore, to facilitate diffusion processing, we introduce a temporal spike encoding block and a decoding mechanism to convert static inputs into temporal spike signals and revert the output temporal spikes into static noise. A significant challenge arises from manually setting membrane thresholds for LIF neurons, which can compromise SDP performance and necessitate extensive training periods. We identify pronounced disparities in membrane potential accumulation speeds and spiking fire rates across channels. Addressing these issues, we propose Learnable Channel-wise Membrane Thresholds (LCMT), enabling adaptive acquisition of membrane potential thresholds during training via back-propagation, while keeping the low extra computation cost by sharing the thresholds across channels. Evaluating the SDP model on seven distinct tasks with SNN timestep $T_S=4$, we achieve results comparable to those of the ANN counterparts, along with faster convergence speeds than the baseline SNN method. This improvement is accompanied by a reduction of 94.3\% in dynamic energy consumption estimated on 45nm hardware.

        
        
    \section{Related works}
\label{sec:related}

\subsection{Spiking Neural Networks}


\textbf{Spiking neural network in robotics:}
In robotics, the proposed multi-task autonomous learning paradigm~\cite{liu2021spiking} for mobile robots employs an SNN controller with a reward-modulated Spiking-time-dependent Plasticity learning rule and a task switch mechanism to enable the robot to autonomously learn, switch, and complete obstacle avoidance and target tracking tasks. \cite{jiang2023fully} presents the successful application of a novel Spiking Neural Network (SNN) to control legged robots, which achieves outstanding results across simulated terrains while offering advantages in inference speed, energy consumption, and biological interpretability over traditional artificial neural networks. In robotic arm manipulation, \cite{marrero2024novel} presents the design and analysis of a new PID controller based on the spiking neural network (SNN) for a 3-DoF robotic arm, which demonstrates improved accuracy and efficiency in trajectory tracking compared to conventional neural network and fuzzy controllers.

\textbf{Learnable membrane parameters in SNNs:}
Some papers focus on developing bioinspired and adaptive membrane parameters for SNNs to improve their performance and efficiency.
Ding et al.~\cite{ding2022biologically} proposes a bioinspired dynamic energy-temporal threshold (BDETT) scheme that mirrors the biological observation of a dynamic threshold positively correlated with average membrane potential and negatively correlated with the preceding rate of depolarization.
Wei et al.~\cite{wei2023temporal} examines the limitations of applying existing time-to-first-spike (TTFS) learning algorithms and introduces a dynamic firing threshold (DFT) mechanism and a novel direct training algorithm for TTFS-based deep SNNs called DTA-TTFS.
Fang et al.~\cite{fang2021incorporating} proposes a training algorithm that can learn the synaptic weights and membrane time constants of SNNs simultaneously, inspired by the observation that membrane-related parameters differ across brain regions. 
The proposed methods in this work share similarities with the learnable thresholding mechanism introduced by Wang et al.~\cite{wang2022ltmd}. However, the study in this paper employs a channel-wise adaptive membrane potential threshold, which differs from the prior approach.

\subsection{Diffusion Policy}



Currently, diffusion models\cite{sohl2015deep} and DDPMs~\cite{ho2020denoising} have been widely used in many related fields, especially in the field of computer vision, such as high-resolution image generation~\cite{dhariwal2021diffusion,rombach2022high}, image restoration~\cite{li2023diffusion}, super-resolution tasks~\cite{shang2024resdiff,wu2023super}, text-to-image generation~\cite{gu2022vector,kim2022diffusionclip,nichol2021glide}, and other scenarios~\cite{okhotin2024star,wu2024medsegdiff}. The diffusion algorithm most relevant to this work is the Diffusion Policy algorithm designed for robotic manipulation proposed by Chi et al.~\cite{chi2023diffusion}. They built a robot's visuomotor policy as a conditional denoising diffusion process. This approach evaluates the Diffusion Policy across 12 different tasks from 4 different robot manipulation benchmarks.

    \section{Background}
\label{sec:background}

\subsection{Leaky Integrate-and-Fire Node}
\label{subsec:snn_lif}


    While elaborate conductance-based neuron models can accurately replicate electrophysiological measurements, their complexity makes them challenging to analyze in depth. Therefore, in the context of Spiking Neural Networks research, the most commonly employed neural model is the simplified Leaky Integrate-and-Fire (LIF) model.
    
    The LIF model is established as an RC-circuit composed of a resistor $R$ and a capacitor $C$ in parallel. Its dynamics are defined by two essential components: the evolution of the membrane potential $u(t)$ and a mechanism to generate spikes. According to elementary laws from the theory of electricity, the evolution of the membrane potential is described by the following equation:Eq.~\ref{equ:RC_circuit}:
    \begin{equation}
    \tau_m \frac{du}{dt} = -[u(t) - u_{rest}] + RI(t),
    \label{equ:RC_circuit}
    \end{equation}
    where the input current $I(t)$ can be a continuous current or a short current pulse. We refer to $u$ as the membrane potential and to $\tau_m=RC$ as the membrane time constant of the neuron.
    Spikes are generated to transmit to postsynapse targets whenever the membrane potential $u$ crosses a threshold $\theta$ from below, as described by Eq.~\ref{equ:cross_theta}:
    \begin{equation}
    S(t) = \mathcal{H}(u(t) - \theta),
    \label{equ:cross_theta}
    \end{equation}
    where $\mathcal{H}$ is the Heaviside step function.

\subsection{Diffusion Model}




The training process of DDPMs is designed to minimize the difference between the forward diffusion process and the reverse denoising process. Specifically, the training goal is to minimize reconstruction errors by maximizing logarithmic likelihood. This is usually done by the Variational inference method, which defines a Variational Lower Bound (VLB):
\begin{equation}
\mathcal{L} = \mathbb{E}_q \left[ \sum_{t=1}^{T} D_{KL}\left(q(x_{t-1} \mid x_t, x_0) \parallel p_w(x_{t-1} \mid x_t)\right) \right]
\end{equation}
where \( D_{KL} \) is Kullback-Leibler divergence, which measures the difference between the real distribution and the model distribution.
    \vspace{-3mm}
\section{Method}
\label{sec:method}

	\begin{figure}[t]
		\centering
			\includegraphics[width=1\linewidth]{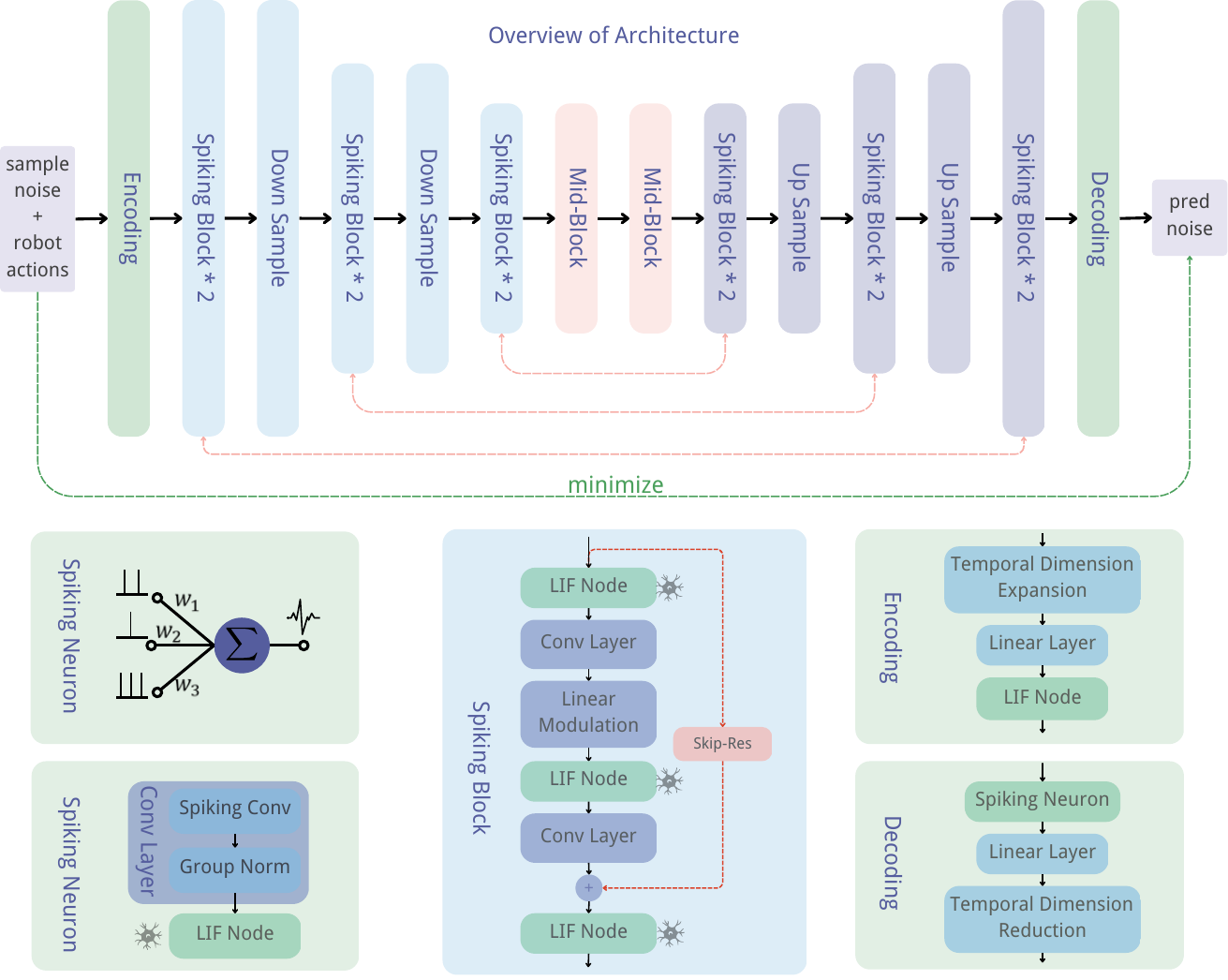}
		\caption{Illustration of architecture and sub-modules of the proposed SDP model. The top of the figure illustrates the overall architecture of the model, which is a Spiking U-Net structure incorporating neuromorphic computing. The model utilizes the diffusion algorithm to iteratively minimize the error between sampling noise and prediction noise in training network parameters. The bottom of the figure showcases the primary sub-modules within the Spiking U-Net, along with their respective operations and neuron connection schemes.}
		\label{fig:arch}
	\end{figure}


\subsection{Overview of SDP Architecture}
\label{subsec:overview_arch}

The overall architecture of the model, as shown in the upper part of Figure 2, is a Spiking U-Net. The input, sampled from random Gaussian noise, is encoded into a spike sequence of fixed time length $T_{S}$ through an encoding layer. This spike sequence consists of binary values, 0 and 1, representing the refractory and excited states, respectively. The encoded input is fed through consecutive levels of Spiking Blocks and downsampling modules, and then upsampled through multiple layers to restore to the input data dimensions. To evaluate the error between prediction noise and sampling noise, the dynamic spike signals formed by the temporal sequence need to undergo a decoding layer to be transformed back into static noise output. Each spiking block comprises two spiking neurons that form a residual connection block, with each spiking neuron composed of a convolutional layer and Leaky Integrated and Fire neurons. The neuron performs convolution operations (i.e., summation operations) on the spike state sequence received from the upper neurons within the time $T_{S}$ at time $t_s$, accumulating the operation results into the current neuron's membrane potential. Once the accumulated membrane potential exceeds the set threshold at time $t_s$, the neuron emits a spike to the next layer neuron at time $t_s$, as described in Sec.~\ref{subsec:snn_lif}. 

	\begin{figure}[t]
		\centering
		\subfigure[]{
			\begin{minipage}[t]{0.23\linewidth}
				\centering
				\includegraphics[width=1\linewidth]{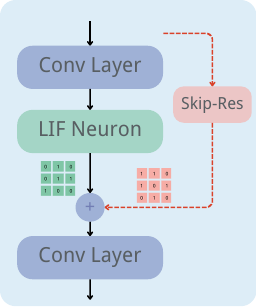}
                \label{fig:skip_a}
			\end{minipage}%
		}%
		\subfigure[]{
			\begin{minipage}[t]{0.23\linewidth}
				\centering
				\includegraphics[width=1\linewidth]{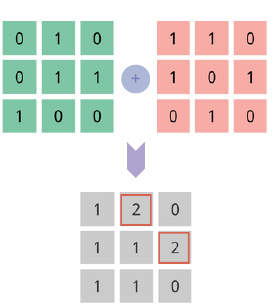}
                \label{fig:skip_b}
			\end{minipage}%
		}
        \subfigure[]{
			\begin{minipage}[t]{0.23\linewidth}
				\centering
				\includegraphics[width=1\linewidth]{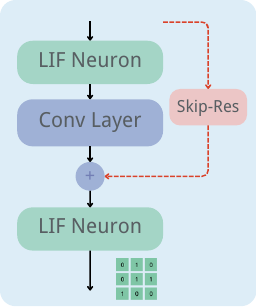}
                \label{fig:skip_c}
			\end{minipage}%
		}%
		\centering
		\caption{Skip residual connection.}
		\label{fig:skip}
	\end{figure}

    Common residual connections are typically established after a non-linear activation function as illustrated in Fig.~\ref{fig:skip_a}, connecting the current activation with the previous activation, and performing summation operations. In this study, the LIF node serves as a non-linear unit emitting spike signals. If following the conventional approach and placing the residual connection after the LIF node, the 0-1 state emitted by the current LIF node would be summed with the 0-1 state of the previous spike data. The 0-1 values in spike data do not represent actual numerical values but rather two relative states. Summing these two states at the position of the residual connection in spike data would result in ternary states, disrupting the neural morphology of the network, as illustrated in Fig.~\ref{fig:skip_b}. In this paper, we position the residual connections of neurons after the conv layer. The data after spiking conv operations itself consists of floating-point data, enabling summation without any cost. The resulting sum is then passed into the LIF node, generating spike data that can propagate normally to subsequent neurons as shown in Fig.~\ref{fig:skip_c}.
    
\vspace{-1.5mm}
\subsection{Temporal Spiking Encoding And Decoding}
\label{subsec:encode_decode}


    Research on spiking neural networks is in its early stages, with many designs drawing inspiration from traditional artificial neural networks (ANNs), thus retaining aspects of ANN architecture. Traditional ANNs operate as parallel processing networks, where static data generated at a single moment is synchronously fed into the network. In contrast, SNNs represent a dynamic temporal network structure, requiring input data that is sequential in the time dimension. When attempting to apply SNN network structures to address problems traditionally solved by ANN networks, a challenge arises in the conversion and migration between static and dynamic data formats. To integrate datasets born from the foundation of traditional ANN architectures into SNNs, temporal spike encoding of the data is necessary, along with the conversion of temporal spike sequences output by SNN networks back into static data to achieve the desired task.

\begin{figure}[t]
		\centering
		\subfigure[]{
			\begin{minipage}[t]{0.85\linewidth}
				\centering
				\includegraphics[width=1\linewidth]{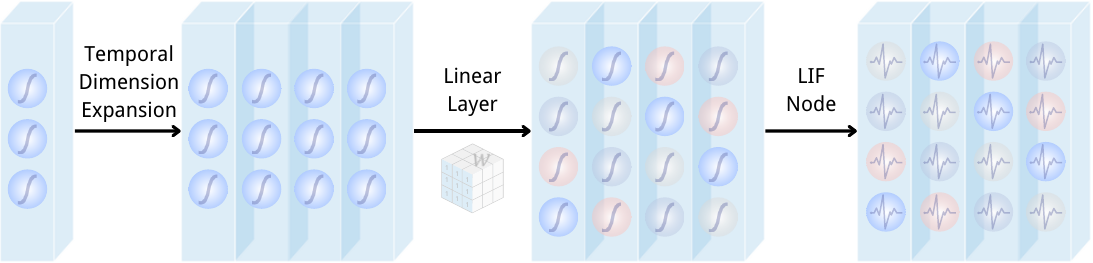}
			\end{minipage}%
		}%
        \\
		\subfigure[]{
			\begin{minipage}[t]{0.85\linewidth}
				\centering
				\includegraphics[width=1\linewidth]{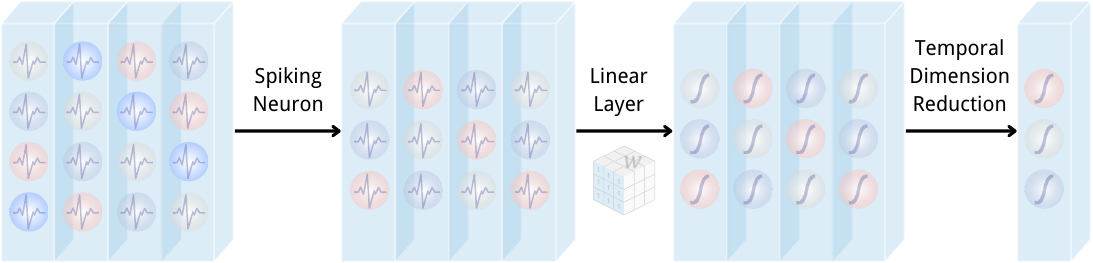}
			\end{minipage}%
		}
		\centering
		\caption{Temporal spiking (a) encoding and (b) decoding}
		\label{fig:coding}
	\end{figure}

    In this study, we introduce a temporal spike encoding block to transform the network's static input data into pulse signals, as well as a Temporal compression decoding block to convert the temporal spikes output by the network back into static noise estimates, as shown in Fig.~\ref{fig:coding}.

\subsection{Learnable Channel-wise Membrane Threshold}
\label{subsec:LCMT}

    The spiking neuron, as the most fundamental and crucial component of Spiking Neural Networks (SNNs), processes spike signals received from presynaptic neurons. The result of this information processing is reflected in the neuron's membrane potential. The evolution process of the spiking neuron's membrane potential is described by Eq.~\ref{eq:evo_mem}. In this equation, $u_i^n[t]$ represents the membrane potential of the $i$-$th$ neuron in the $n$-$th$ layer at time $t$. The term $s_i^n[t-1]$ indicates the spike state emitted by the $i$-$th$ neuron in the $n$-$th$ layer at time $t-1$. Here, $\tau$ denotes the time decay constant, $\delta(i,j)$ represents the connectivity between neurons $i$ and $j$ in different layers, and $W_{ij}$ signifies the weight of the connection between the neuron $i$ and the neuron $j$. 

    \begin{equation}
        u^n_i[t] = \tau u_i^n[t-1](1-s_i^n[t-1]) + \sum_{j\in \delta(i,j)}W_{ij}s_j^{n-1}[t]
    \label{eq:evo_mem}
    \end{equation}

    Once the accumulated membrane potential surpasses a predetermined threshold, a spike signal is emitted to postsynaptic neurons. The state equation for spike emission in a spiking neuron is given by Eq.~\ref{eq:spike}:
    \begin{equation}
        s^n_i[t] = \mathcal{H}[u_i^n[t]-\theta_i^n]
    \label{eq:spike}
    \end{equation}
    where $\mathcal{H}$ represents the Heaviside step function and $\theta_i^n$ denotes the membrane potential threshold for the $i$-$th$ neuron in the $n$-$th$ layer.

    The membrane threshold in a spiking neuron is a crucial hyperparameter to determine both the firing rates and the momentary amplitude of the membrane potential at the firing time. However, as a hyperparameter, a manually set threshold may not always align well with network parameters. For instance, a high threshold can make it difficult for the membrane potential to accumulate to a sufficient level to fire a spike, while a low threshold can cause neurons to fire spikes too easily, leading to reduced information content and increased power consumption. Furthermore, repeatedly trying different thresholds during training would significantly increase the model's training costs. Consequently, this issue motivates the introduction of a learnable membrane potential threshold, which allows neurons to autonomously adjust and self-shape.

    Introducing a learnable membrane potential threshold involves treating the threshold of each spiking neuron as a trainable parameter. As depicted in Eq.~\ref{eq:partial_theta}, this approach enables the computation of threshold gradients during backpropagation, thereby allowing the optimization of threshold values through parameter updates. 
    \begin{equation}
        \frac{\partial \mathcal{L}}{\partial\theta_i^n} = \sum_{t=1}^{T_{S}}\frac{\partial\mathcal{L}}{\partial s_i^n} \frac{\partial s_i^n}{\partial \theta_i^n}
    \label{eq:partial_theta}
    \end{equation}

    Unfortunately, the Heaviside step function is discontinuous and has a derivative of zero everywhere except at the point of discontinuity. Therefore, when using the chain rule to compute the gradient of the Heaviside function during backpropagation, a surrogate gradient must be employed, as described in Eq.~\ref{eq:surrogate}.
    \begin{equation}
      \mathcal{S}^{'}[\mathcal{H}[x]] = \left\{\begin{array}{ccl}
            1 & \mbox{for}
            & 0 \le x \le 0.5 \\ 0 & \mbox{for} & otherwise \\
                \end{array} \right.
      \label{eq:surrogate}
    \end{equation}

	\begin{figure}[t]
		\centering
		\subfigure[]{
			\begin{minipage}[t]{0.5\linewidth}
				\centering
				\includegraphics[width=1\linewidth]{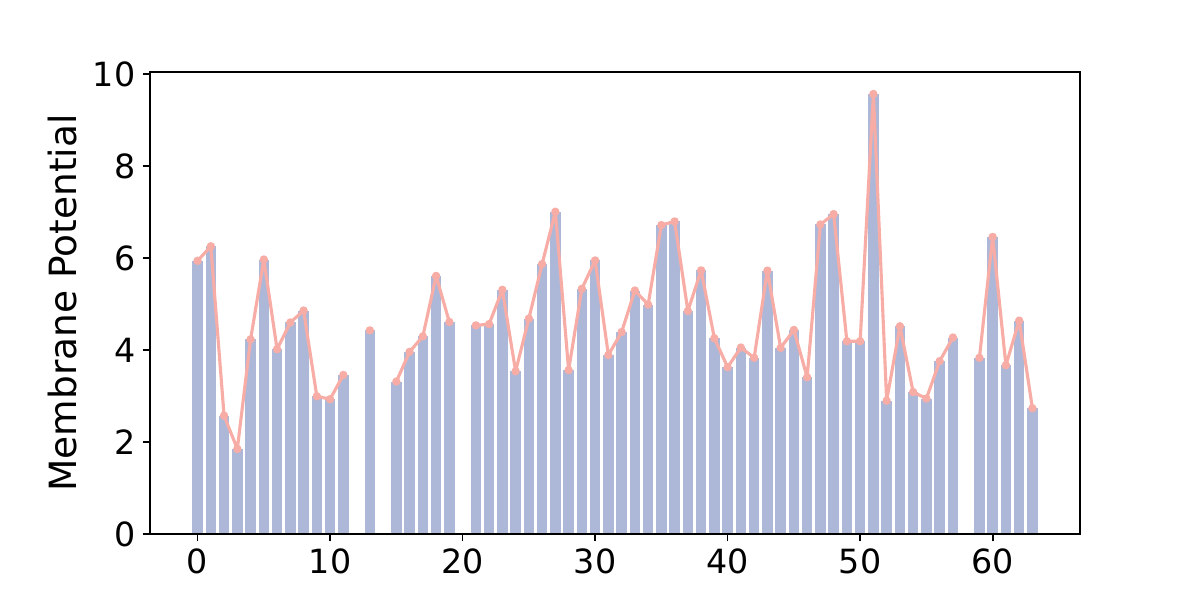}
			\end{minipage}%
		}
		\subfigure[]{
			\begin{minipage}[t]{0.5\linewidth}
				\centering
				\includegraphics[width=1\linewidth]{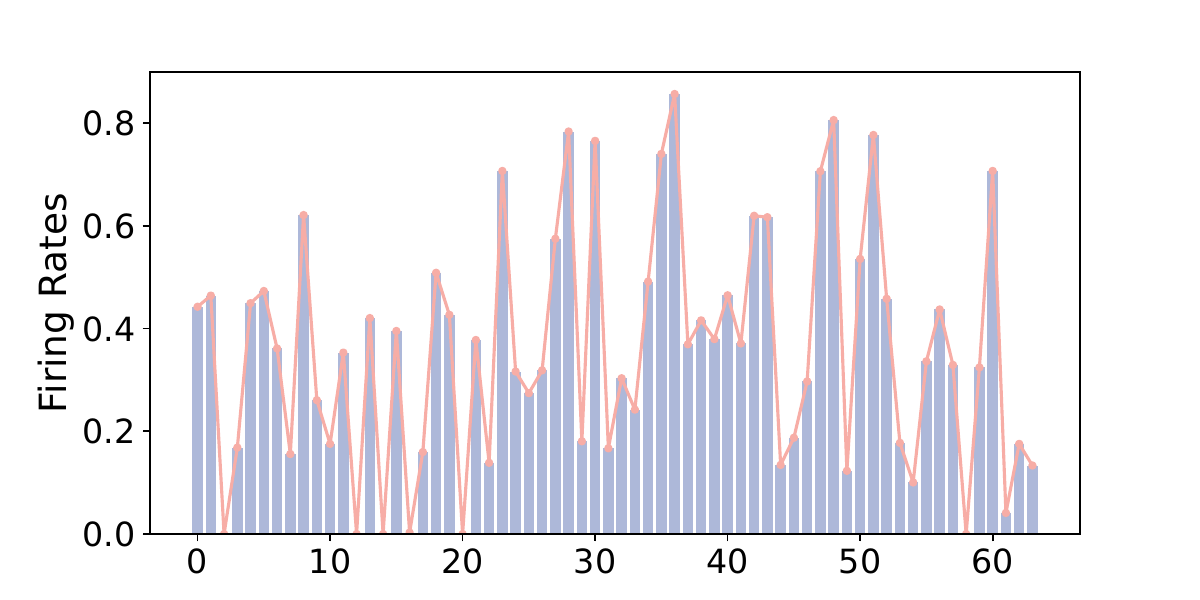}
			\end{minipage}%
		}%
		\centering
		\caption{
        (a) Momentary values of the membrane potential at the firing time selected from the first 64 channels. 
        (b) Enhancing the illustration of discrepancies in spike firing rates across the selected first 64 channels.}
		\label{fig:diff_ch}
	\end{figure}

	\begin{table*}[!t]
    	\caption{Experimental Results.} 
		\label{tab:main-result}
		\centering
		\begin{tabular}{c|c|c|c|c|c|c|c|c|c}  
			\hline       
			\hline
			\diagbox{Methods}{Tasks} & \multirow{2}{4em}{\centering SNN?}&\multirow{2}{3em}{\centering Push-T} &\multicolumn{2}{c|}{BlockPush}& \multicolumn{5}{c}{RoboMimic} \\
			\cline{4-5} \cline{6-10} 
			&&&@p1&@p2& Lift & Can & Square & ToolHang & Transport  \\  
			\hline
			LSTM-GMM~\cite{mandlekar2021matters} &\ding{55}&0.67&0.03&0.01&\underline{1.00}&\underline{1.00}&0.95&\underline{0.67}&0.76\\
            IBC~\cite{florence2022implicit} &\ding{55}&0.90&0.01&0.00&0.79&0.00&0.00&0.00&0.00\\
            BET~\cite{shafiullah2022behavior} &\ding{55}&0.79&\underline{0.96}&\underline{0.71}&\underline{1.00}&\underline{1.00}&0.76&0.58&0.38\\
            Diffusion-Policy~\cite{chi2023diffusion} &\ding{55}&\underline{0.95}&0.36&0.11&\underline{1.00}&\underline{1.00}&\underline{1.00}&0.50&\underline{0.94}\\
            \hline
            SDDPM~\cite{cao2024spiking} &\ding{52}&0.86&\textbf{0.22}&\textbf{0.06}&0.98&0.82&0.82&0.02&0.00\\
            \textbf{SDP(Ours)} &\ding{52}&\textbf{0.92}&0.12&0.02&\textbf{1.00}&\textbf{0.98}&\textbf{0.96}&\textbf{0.52}&\textbf{0.82}\\
            \hline
		\end{tabular}
	\end{table*}

    Individually learning a large number of neuronal membrane potential thresholds undoubtedly increases computational complexity and the number of model parameters. Upon further investigation, we observed that there are noticeable differences in spike firing rates and the momentary membrane potential at firing time between neurons in the same layer but across different channels. However, these variables remain relatively consistent within the same channel, as shown in Fig.~\ref{fig:diff_ch}. Building upon the foundation of learnable membrane potential thresholds, we introduced the concept of Learnable Channel-wise Membrane Threshold (LCMT) to adapt to the varying membrane potential accumulation states and spike firing rates across different channels. LCMT relaxes the constraints on membrane potential thresholds, such that spiking neurons within the same channel utilize a consistent membrane potential threshold $\theta^c$, as illustrated in Eq.~\ref{eq:partial_theta_ch}. For brevity, we have omitted the superscript $n$.
    \begin{equation}
        \frac{\partial \mathcal{L}}{\partial\theta^c} = \sum_{t=1}^{T_{S}}\frac{\partial\mathcal{L}}{\partial s_i^c} \frac{\partial s_i^c}{\partial \theta^c}
    \label{eq:partial_theta_ch}
    \end{equation}

    \begin{figure}[t]
      \centering
       \includegraphics[width=0.7\linewidth]{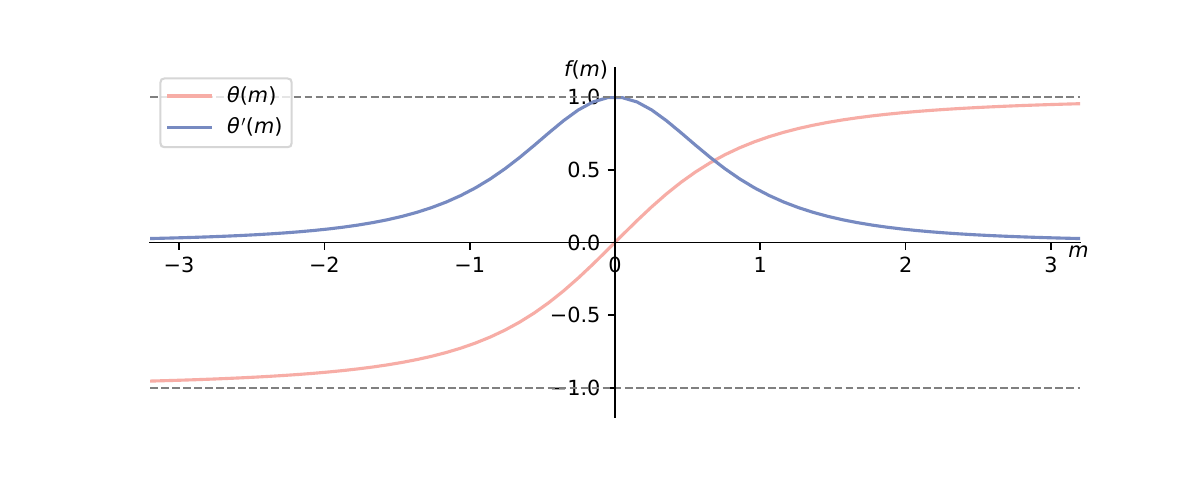}
       \caption{Illustration of membrane threshold and its derivatives.}
       \label{fig:curve}
    \end{figure}

    In the absence of constraints, the range for the membrane potential threshold $\theta^c$ is $(-\infty,+\infty)$. However, excessively large or small values for this threshold are undesirable. Consequently, it is preferable to confine the threshold to a specific narrow range without introducing additional constraints. Inspired by \cite{wang2022ltmd}, we introduce a new parameter $m^c$ with a range of $(-\infty,+\infty)$ to limit the range of $\theta^c$ to a range of $(-1,1)$ effectively, as illustrated in Fig.~\ref{fig:curve}. By setting  
    \begin{equation}
        \theta = \frac{m}{\sqrt{1+m^2}},
    \label{eq:sur_theta}
    \end{equation}
    The derivative of Eq.~\ref{eq:sur_theta} is expressed as: 
    \begin{equation}
        \frac{\partial \theta^c}{\partial m^c} = \frac{1}{(1+(m^c)^2)(\sqrt{1+(m^c)^2})}.
    \label{eq:partial_theta_to_m}
    \end{equation}
    Given these derivations, substituting Eq.~\ref{eq:surrogate},~\ref{eq:sur_theta}, and~\ref{eq:partial_theta_to_m} into Eq.~\ref{eq:partial_theta_ch} yields the gradient formula for the learnable channel-wise membrane threshold as expressed Eq.~\ref{eq:partial_m}. 
    \begin{equation}
        \begin{aligned}
        \frac{\partial \mathcal{L}}{\partial m^c} &= \sum_{t=1}^{T_{S}}\frac{\partial\mathcal{L}}{\partial s_i^c} \frac{\partial s_i^c}{\partial \theta_i^c}\frac{\partial \theta_i^c}{\partial m^c} \\
        &=\sum_{t=1}^{T_{S}}\frac{\partial\mathcal{L}}{\partial s_i^c} \mathcal{S}^{'}[\mathcal{H}[u_i^c[t]-\frac{m^c}{\sqrt{1+(m^c)^2}}]] \\
        &~~~~\frac{1}{\left(1+(m^c)^2\right)\left(\sqrt{1+(m^c)^2}\right)}
        \end{aligned}
        \label{eq:partial_m}
    \end{equation}
    During backpropagation, this gradient is utilized to update the thresholds.

    \section{Experiments}
\label{sec:exp}

	\begin{figure}[t]
		\centering
			\includegraphics[width=1\linewidth]{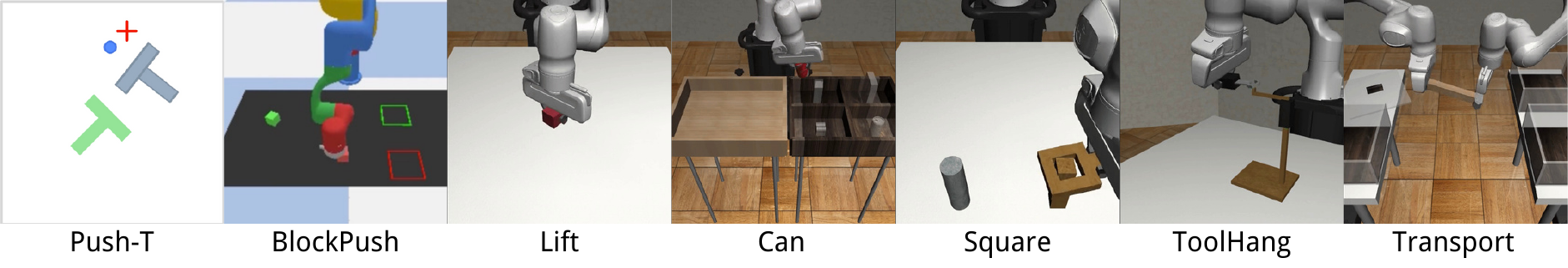}
		\caption{Evaluation tasks.}
		\label{fig:tasks}
	\end{figure}

	\begin{table*}[t]
    	\caption{Comparison of Threshold Guidance with SDDPM.} 
		\label{tab:compare_SDDPM}
		\centering
        \scalebox{1}{
		\begin{tabular}{c|c|c|c|c|c|c|c|c|c}  
			\hline       
			\hline
			\multirow{2}{6em}{\centering Methods}
            & \multirow{2}{4em}{\centering Threshold}&\multirow{2}{3em}{\centering Push-T} &\multicolumn{2}{c|}{BlockPush} & \multicolumn{5}{c}{RoboMimic} \\
			\cline{4-5} \cline{6-10} 
			&&&@p1&@p2& Lift & Can & Square & ToolHang & Transport  \\  
			\hline
            \multirow{6}{6em}{\centering SDDPM \cite{cao2024spiking}} 
            &$-0.001$&0.79&0.10&0.00&0.86&0.72&0.82&0.00&0.00\\
            &$-0.002$&0.79&\textbf{0.22}&\textbf{0.06}&0.92&0.74&0.74&0.00&0.00\\
            &$-0.003$&0.80&0.22&0.00&0.96&0.76&0.64&0.00&0.00\\
            &$+0.001$&0.82&0.20&0.06&0.90&0.72&0.78&0.00&0.00\\
            &$+0.002$&0.80&0.20&0.00&0.92&0.72&0.72&0.00&0.00\\
            &$+0.003$&0.82&0.06&0.00&0.92&0.72&0.68&0.00&0.00\\
            \hline
            \textbf{SDP(Ours)} &\textbf{LCMT}&\textbf{0.92}&0.12&0.02&\textbf{1.00}&\textbf{0.98}&\textbf{0.96}&\textbf{0.52}&\textbf{0.82}\\
			\hline
		\end{tabular}
        }
	\end{table*}

	\begin{table}[t]
    	\caption{Ablation Studies for SNN Timesteps $T_{S}$.} 
		\label{tab:ablation_T}
		\centering
        \scalebox{0.78}{
		\begin{tabular}{c|c|c|c|c|c|c|c|c}  
			\hline       
			\hline
			\multirow{2}{3em}{\centering Methods}&\multirow{2}{3em}{\centering Push-T} &\multicolumn{2}{c|}{BlockPush} & \multicolumn{5}{c}{RoboMimic} \\
			\cline{3-4}\cline{5-9} 
			&&@p1&@p2& Lift & Can & Square & ToolHang & Transport  \\  
			\hline
			$T_{S}=1$ &0.90&0.18&0.06&\textbf{1.00}&0.96&0.94&0.26&0.26\\
            $T_{S}=2$ &0.90&0.08&0.00&\textbf{1.00}&0.96&0.92&0.40&0.86\\
            $T_{S}=4$ &\textbf{0.92}&0.12&0.02&\textbf{1.00}&0.98&\textbf{0.96}&\textbf{0.52}&0.82\\
            $T_{S}=8$ &\textbf{0.92}&\textbf{0.22}&\textbf{0.06}&\textbf{1.00}&\textbf{1.00}&0.92&0.48&\textbf{0.88}\\
			\hline
		\end{tabular}
        }
	\end{table}

    \begin{table}[t]
    	\caption{Ablation Studies for LCMT.} 
		\label{tab:ablation_LCMT}
		\centering
        \scalebox{0.75}{
		\begin{tabular}{c|c|c|c|c|c|c|c|c}  
			\hline       
			\hline
			\multirow{2}{2.8em}{\centering Methods} &\multirow{2}{3em}{\centering Push-T} &\multicolumn{2}{c|}{BlockPush} & \multicolumn{5}{c}{RoboMimic} \\
			\cline{3-4}\cline{5-9} 
			&&@p1&@p2& Lift & Can & Square & ToolHang & Transport  \\  
			\hline
		  w/o LCMT &0.86&\textbf{0.20}&\textbf{0.06}&\textbf{1.00}&0.82&0.80&0.02&0.00\\
            + LCMT &\textbf{0.90}&0.08&0.00&\textbf{1.00}&\textbf{0.98}&\textbf{0.96}&\textbf{0.52}&\textbf{0.86}\\
			\hline
		\end{tabular}
        }
	\end{table}

\subsection{Datasets and Experimental Settings}
The SDP model proposed in this study is evaluated in seven robotic manipulation tasks, following the experimental setup and evaluation metrics used by Chi et al.~\cite{chi2023diffusion}. These tasks include \textbf{\textit{Push-T}}~\cite{florence2022implicit}, \textit{\textbf{BlockPush}}~\cite{shafiullah2022behavior}, \textit{\textbf{Lift}}, \textit{\textbf{Can}}, \textit{\textbf{Square}}, \textit{\textbf{ToolHang}}, and \textit{\textbf{Transport}}. They encompass a diverse range of actions, such as pushing, grasping, and lifting, as well as interactions between robotic arms and their environment, as illustrated in Fig.~\ref{fig:tasks}. To elaborate, the \textit{\textbf{Push-T}} task requires using a circular end-effector to push a T-shaped object into a fixed target box. The objective of \textit{\textbf{BlockPush}} is to push two blocks into two different squares in any order. Both of these tasks are performed in a 2D tabletop environment. The remaining five tasks are sourced from \textit{\textbf{RoboMimic}}~\cite{mandlekar2021matters} and are conducted in a 3D spatial setting, utilizing datasets collected from proficient human teleoperated demonstrations. Specifically:
\begin{itemize}
    \item \textit{\textbf{Lift}}: Involves picking up a block.
    \item \textit{\textbf{Can}}: Focuses on grasping and moving a soda can.
    \item \textit{\textbf{Square}}: Requires lifting an object with a square hole and fitting it onto a square pillar.
    \item \textit{\textbf{ToolHang}}: Involves grabbing a tool and suspending it.
    \item \textit{\textbf{Transport}}: Involves two robotic arms that pass an object between them.
\end{itemize}
This diverse set of tasks provides a comprehensive evaluation of the capabilities of the proposed SDP model. In terms of additional experimental settings, the timestep $T_S$ for the spiking neural network is set to 4, while the timestep $T_D$ for the diffusion process is set to 100. The proposed model is evaluated on a single NVIDIA RTX 4090 GPU to ensure sufficient computational performance and reliability.

\subsection{Experimental Results}
During the training process, checkpoints are saved every 50 epochs. The reported results are based on the average evaluation metrics of the best-performing checkpoint. For most tasks, we use the success rate as the evaluation metric; however, for the push-T task, we use target area coverage. We compare two categories of methods: one using artificial neural networks (ANNs), including LSTM-GMM~\cite{mandlekar2021matters}, IBC~\cite{florence2022implicit}, BET~\cite{shafiullah2022behavior}, and Diffusion Policy~\cite{chi2023diffusion}; and another using spiking neural networks (SNNs). We used SDDPM~\cite{cao2024spiking} as the baseline SNN in our diffusion policy. For each baseline method, we selected their best performance across various benchmarks. Specifically, for the SDDPM baseline, we detailed the results obtained by adjusting all membrane thresholds. For the BlockPush task, $p_x$ represents the frequency of pushing $x$ blocks into the target area. The results from all these baseline methods and our proposed SDP model are summarized in Tab.~\ref{tab:main-result} and Tab.~\ref{tab:compare_SDDPM}.

\subsection{Ablation Studies}

\subsubsection{Ablation for Timestep $T_{S}$}
For the ablation study on the SNN timestep $T_S$, we set 
$T_S$ to 1, 2, 4, and 8 and report the SDP model's performance on each benchmark, as shown in Tab.~\ref{tab:ablation_T}.  The ablation study on the SNN timestep demonstrates that, for most tasks, performance generally improves with an increase in $T_S$. However, once $T_S$ reaches 4, further increases have a negligible impact on the results.  Instead, larger values of $T_S$extend the temporal dimension, thereby increasing latency and power consumption. Consequently, we opted for $T_S=4$ as the optimal trade-off value.

    \subsubsection{Ablation for LCMT}
\begin{figure}[t]
		\centering
		\subfigure[]{
			\begin{minipage}[t]{0.45\linewidth}
				\centering
				\includegraphics[width=1\linewidth]{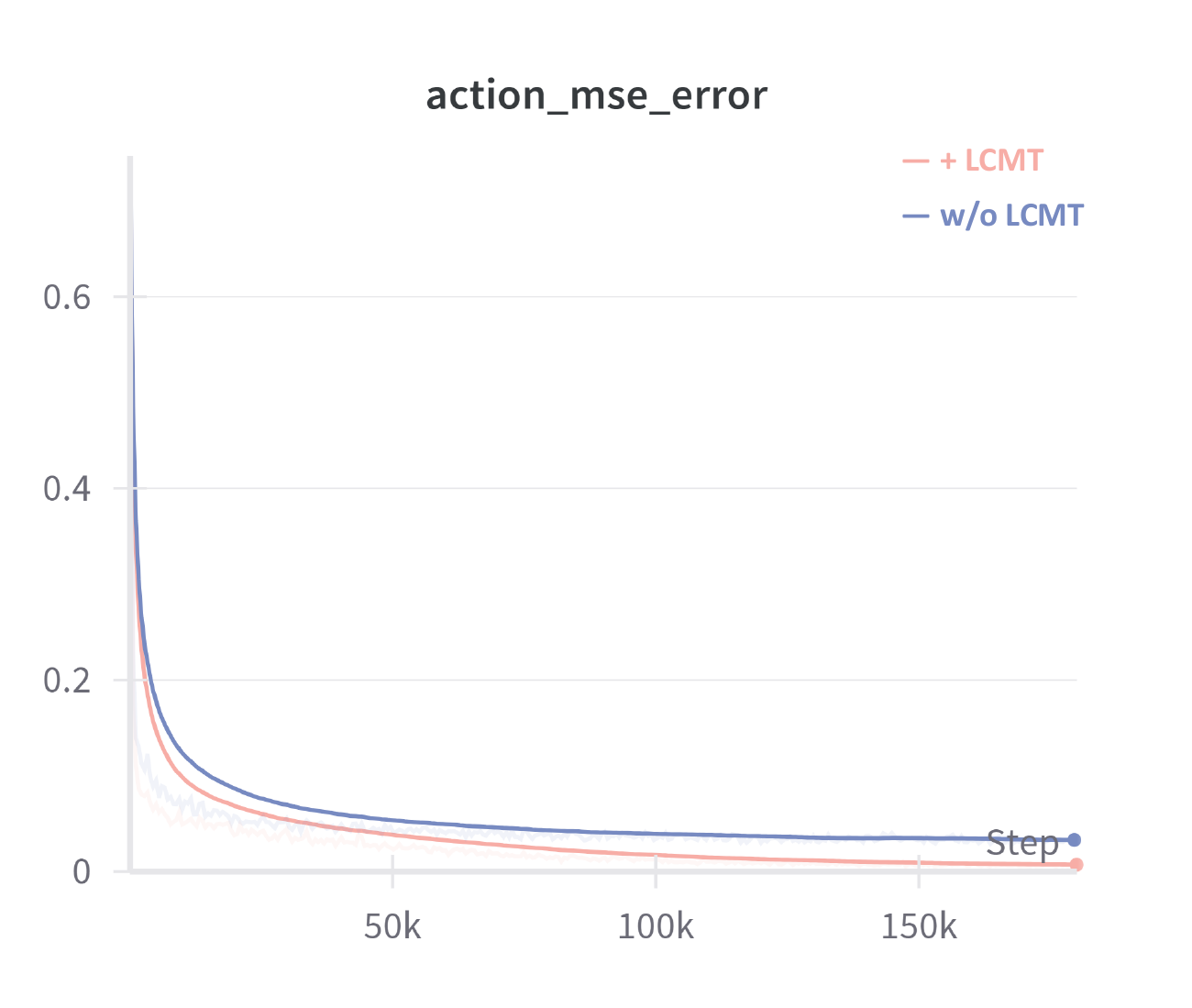}
			\end{minipage}%
		}%
		\subfigure[]{
			\begin{minipage}[t]{0.45\linewidth}
				\centering
				\includegraphics[width=1\linewidth]{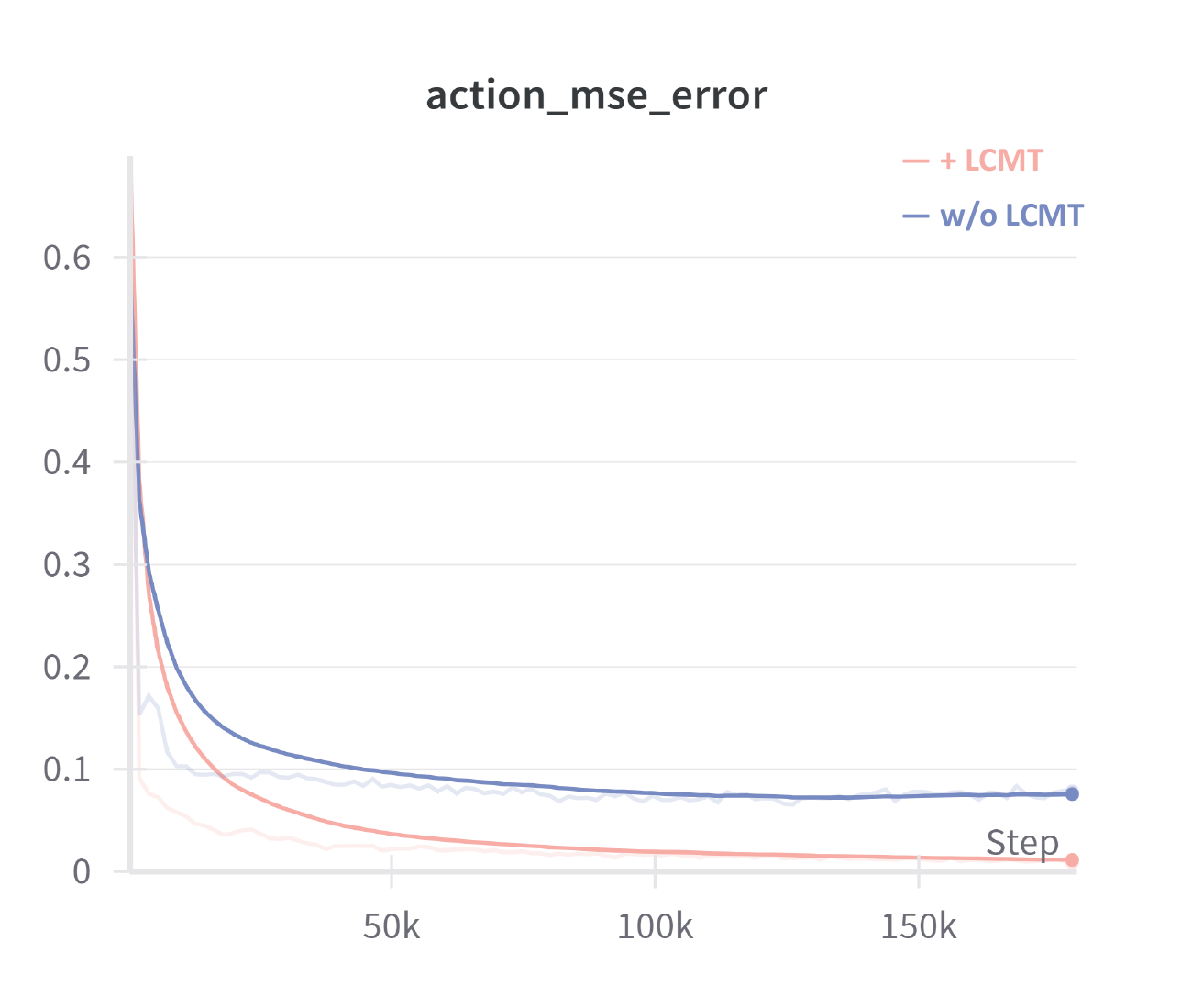}
			\end{minipage}%
		}
		\centering
		\caption{Action MSE error on (a) \textit{\textbf{Square}} and (b) \textit{\textbf{Transport}}.}
		\label{fig:action_mse}
	\end{figure}

    To validate the effectiveness of the proposed LCMT, we conduct ablation studies from two perspectives: \textbf{(1) success rate} on tasks and \textbf{(2) convergence speed}. The results of models with and without LCMT on each benchmark are presented in Tab.~\ref{tab:ablation_LCMT}. The addition of LCMT significantly enhances the model's performance. In terms of convergence speed, we plot the MSE error curves of the robot arm actions, as shown in Fig.~\ref{fig:action_mse} The model equipped with LCMT converges to better results at a faster rate.
    
\subsection{Energy Consumption Analysis}
    The theoretical energy consumption for both SNNs and ANNs, as referenced by~\cite{yao2023attention}, can be calculated as follows:
    \begin{equation}
        E_{SNN} = E_{AC} \times SOPs
        \label{eq:e_snn}
    \end{equation}
    and 
    \begin{equation}
        E_{ANN} = E_{MAC} \times AOPs,
        \label{eq:e_ann}
    \end{equation}
    where $SOPs$ represents the number of spike-based accumulate operations (AC) and $AOPs$ represents the number of multiply-and-accumulate operations (MAC). Assuming that both AC and MAC operations are implemented on 45nm hardware, the energy consumption is $E_{AC}=0.9\mu J$ and $E_{MAC}=4.6\mu J$. The number of AC operations in an SNN is estimated as:
    \begin{equation}
        SOPs = T_S \times \psi \times AOPs,
        \label{eq:sop}
    \end{equation}
    where $\psi$ is the firing rate of the input spike train.

\begin{figure}[!t]
		\centering
		\subfigure[]{
			\begin{minipage}[t]{0.49\linewidth}
				\centering
				\includegraphics[width=1\linewidth]{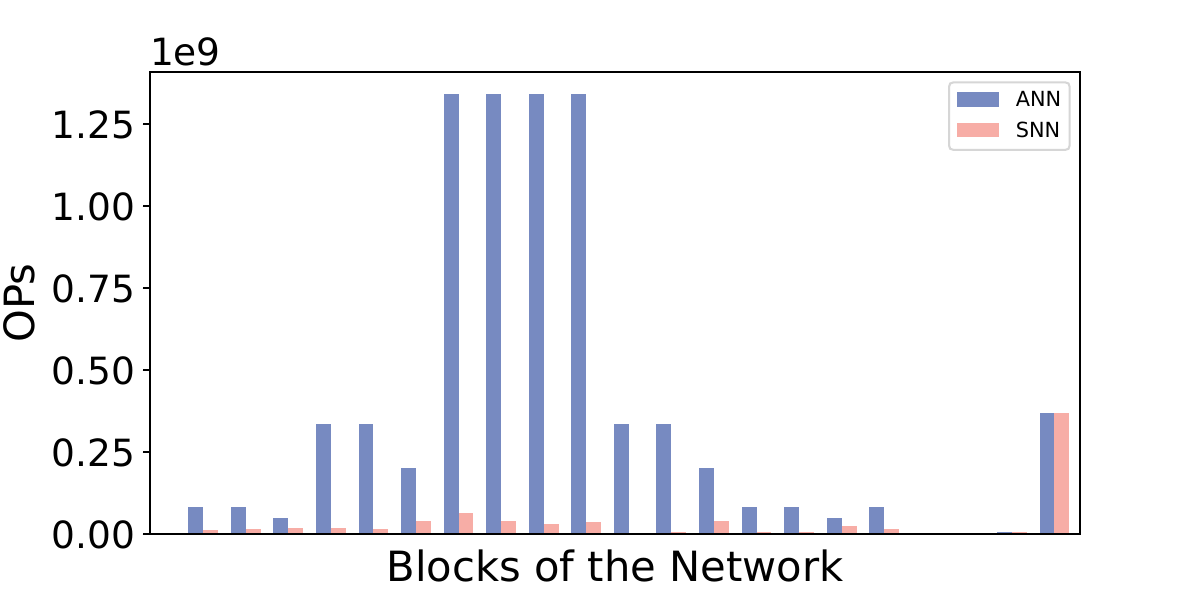}
			\end{minipage}%
		}%
		\subfigure[]{
			\begin{minipage}[t]{0.49\linewidth}
				\centering
				\includegraphics[width=1\linewidth]{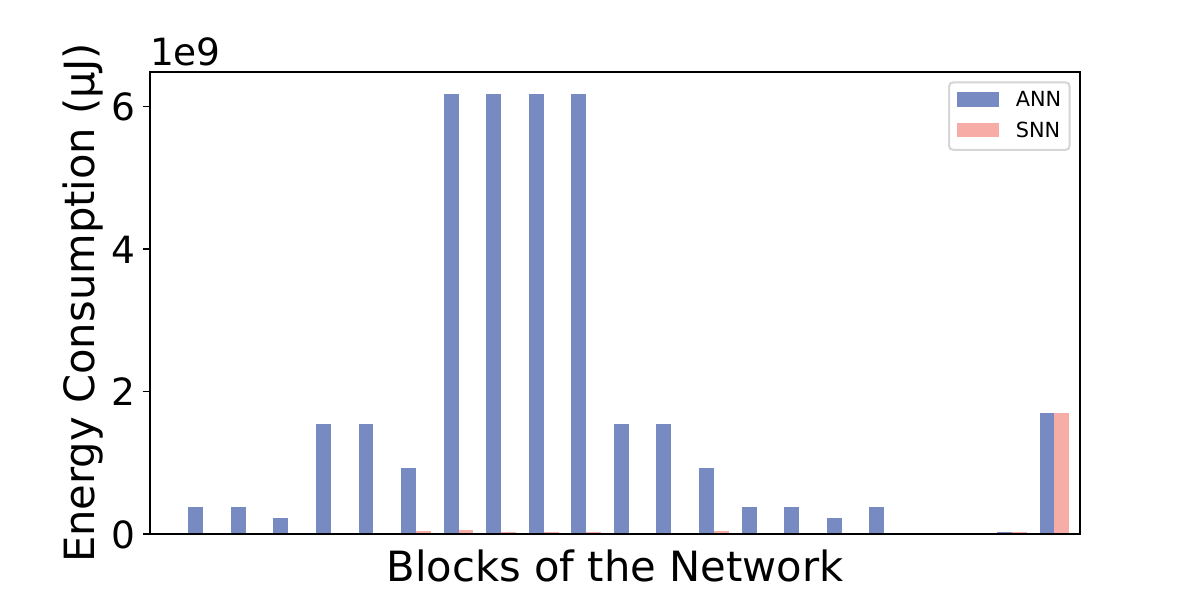}
			\end{minipage}%
		}
		\centering
		\caption{(a) Number of operations. (b) Energy consumption.}
		\label{fig:energy}
	\end{figure}

    Based on the calculations above, the U-Net network that uses spiking neurons reduces 94.3\% of the dynamic energy consumption compared to its ANN counterpart, as shown in Fig.~\ref{fig:energy}. This demonstrates the significant potential of the SDP model proposed in this paper for applications in low-power embodied intelligence.
    
    \section{conclusion}

This study introduces the Spiking Diffusion Policy (SDP) method for robotic manipulation, seamlessly integrating spiking neurons into a diffusion policy framework. This integration enhances computational efficiency while achieving high accuracy in evaluated robotic tasks. We propose the Learning-based Channel Membrane Threshold (LCMT) mechanism to enable the adaptive acquisition of membrane potential thresholds, thereby aligning with the varying membrane potentials and firing rates across channels and eliminating the cumbersome process of manual hyperparameter tuning.
Extensive experiments demonstrate the potential of the SDP model for applications in the field of low-power embodied AI.

    \clearpage 
    \bibliographystyle{IEEEtran}
    \bibliography{example}  

\begin{thebibliography}{10}
\providecommand{\url}[1]{#1}
\csname url@rmstyle\endcsname
\providecommand{\newblock}{\relax}
\providecommand{\bibinfo}[2]{#2}
\providecommand\BIBentrySTDinterwordspacing{\spaceskip=0pt\relax}
\providecommand\BIBentryALTinterwordstretchfactor{4}
\providecommand\BIBentryALTinterwordspacing{\spaceskip=\fontdimen2\font plus
\BIBentryALTinterwordstretchfactor\fontdimen3\font minus \fontdimen4\font\relax}
\providecommand\BIBforeignlanguage[2]{{%
\expandafter\ifx\csname l@#1\endcsname\relax
\typeout{** WARNING: IEEEtran.bst: No hyphenation pattern has been}%
\typeout{** loaded for the language `#1'. Using the pattern for}%
\typeout{** the default language instead.}%
\else
\language=\csname l@#1\endcsname
\fi
#2}}

\bibitem{huang2023voxposer}
W.~Huang, C.~Wang, R.~Zhang, Y.~Li, J.~Wu, and L.~Fei-Fei, ``Voxposer: Composable 3d value maps for robotic manipulation with language models,'' \emph{arXiv preprint arXiv:2307.05973}, 2023.

\bibitem{liu2024robomamba}
J.~Liu, M.~Liu, Z.~Wang, L.~Lee, K.~Zhou, P.~An, S.~Yang, R.~Zhang, Y.~Guo, and S.~Zhang, ``Robomamba: Multimodal state space model for efficient robot reasoning and manipulation,'' \emph{arXiv preprint arXiv:2406.04339}, 2024.

\bibitem{han2015learning}
S.~Han, J.~Pool, J.~Tran, and W.~Dally, ``Learning both weights and connections for efficient neural network,'' \emph{Advances in neural information processing systems}, vol.~28, 2015.

\bibitem{fang2023depgraph}
G.~Fang, X.~Ma, M.~Song, M.~B. Mi, and X.~Wang, ``Depgraph: Towards any structural pruning,'' in \emph{Proceedings of the IEEE/CVF conference on computer vision and pattern recognition}, 2023, pp. 16\,091--16\,101.

\bibitem{sun2019patient}
S.~Sun, Y.~Cheng, Z.~Gan, and J.~Liu, ``Patient knowledge distillation for bert model compression,'' in \emph{Conference on Empirical Methods in Natural Language Processing}, 2019.

\bibitem{li2023q}
X.~Li, Y.~Liu, L.~Lian, H.~Yang, Z.~Dong, D.~Kang, S.~Zhang, and K.~Keutzer, ``Q-diffusion: Quantizing diffusion models,'' in \emph{Proceedings of the IEEE/CVF International Conference on Computer Vision}, 2023, pp. 17\,535--17\,545.

\bibitem{yuan2023rptq}
Z.~Yuan, L.~Niu, J.~Liu, W.~Liu, X.~Wang, Y.~Shang, G.~Sun, Q.~Wu, J.~Wu, and B.~Wu, ``Rptq: Reorder-based post-training quantization for large language models,'' \emph{arXiv preprint arXiv:2304.01089}, 2023.

\bibitem{hou2024fbpt}
Z.~Hou, Y.~Shang, and Y.~Yan, ``Fbpt: A fully binary point transformer,'' \emph{arXiv preprint arXiv:2403.09998}, 2024.

\bibitem{schuman2017survey}
C.~D. Schuman, T.~E. Potok, R.~M. Patton, J.~D. Birdwell, M.~E. Dean, G.~S. Rose, and J.~S. Plank, ``A survey of neuromorphic computing and neural networks in hardware,'' \emph{arXiv preprint arXiv:1705.06963}, 2017.

\bibitem{yu2020overview}
Z.~Yu, A.~M. Abdulghani, A.~Zahid, H.~Heidari, M.~A. Imran, and Q.~H. Abbasi, ``An overview of neuromorphic computing for artificial intelligence enabled hardware-based hopfield neural network,'' \emph{Ieee Access}, vol.~8, pp. 67\,085--67\,099, 2020.

\bibitem{maass1997networks}
W.~Maass, ``Networks of spiking neurons: the third generation of neural network models,'' \emph{Neural networks}, vol.~10, no.~9, pp. 1659--1671, 1997.

\bibitem{diehl2015unsupervised}
P.~U. Diehl and M.~Cook, ``Unsupervised learning of digit recognition using spike-timing-dependent plasticity,'' \emph{Frontiers in computational neuroscience}, vol.~9, p.~99, 2015.

\bibitem{liu2010biologically}
J.~Liu, D.~Perez-Gonzalez, A.~Rees, H.~Erwin, and S.~Wermter, ``A biologically inspired spiking neural network model of the auditory midbrain for sound source localisation,'' \emph{Neurocomputing}, vol.~74, no. 1-3, pp. 129--139, 2010.

\bibitem{fang2021deep}
W.~Fang, Z.~Yu, Y.~Chen, T.~Huang, T.~Masquelier, and Y.~Tian, ``Deep residual learning in spiking neural networks,'' \emph{Advances in Neural Information Processing Systems}, vol.~34, pp. 21\,056--21\,069, 2021.

\bibitem{su2023deep}
Q.~Su, Y.~Chou, Y.~Hu, J.~Li, S.~Mei, Z.~Zhang, and G.~Li, ``Deep directly-trained spiking neural networks for object detection,'' in \emph{Proceedings of the IEEE/CVF International Conference on Computer Vision}, 2023, pp. 6555--6565.

\bibitem{xing2024spikelm}
X.~Xing, Z.~Zhang, Z.~Ni, S.~Xiao, Y.~Ju, S.~Fan, Y.~Wang, J.~Zhang, and G.~Li, ``Spikelm: Towards general spike-driven language modeling via elastic bi-spiking mechanisms,'' \emph{arXiv preprint arXiv:2406.03287}, 2024.

\bibitem{cao2024spiking}
J.~Cao, Z.~Wang, H.~Guo, H.~Cheng, Q.~Zhang, and R.~Xu, ``Spiking denoising diffusion probabilistic models,'' in \emph{Proceedings of the IEEE/CVF Winter Conference on Applications of Computer Vision}, 2024, pp. 4912--4921.

\bibitem{liu2021spiking}
J.~Liu, H.~Lu, Y.~Luo, and S.~Yang, ``Spiking neural network-based multi-task autonomous learning for mobile robots,'' \emph{Engineering Applications of Artificial Intelligence}, vol. 104, p. 104362, 2021.

\bibitem{chi2023diffusion}
C.~Chi, S.~Feng, Y.~Du, Z.~Xu, E.~Cousineau, B.~Burchfiel, and S.~Song, ``Diffusion policy: Visuomotor policy learning via action diffusion,'' \emph{arXiv preprint arXiv:2303.04137}, 2023.

\bibitem{jiang2023fully}
X.~Jiang, Q.~Zhang, J.~Sun, and R.~Xu, ``Fully spiking neural network for legged robots,'' \emph{arXiv preprint arXiv:2310.05022}, 2023.

\bibitem{marrero2024novel}
D.~Marrero, J.~Kern, and C.~Urrea, ``A novel robotic controller using neural engineering framework-based spiking neural networks,'' \emph{Sensors}, vol.~24, no.~2, p. 491, 2024.

\bibitem{ding2022biologically}
J.~Ding, B.~Dong, F.~Heide, Y.~Ding, Y.~Zhou, B.~Yin, and X.~Yang, ``Biologically inspired dynamic thresholds for spiking neural networks,'' \emph{Advances in Neural Information Processing Systems}, vol.~35, pp. 6090--6103, 2022.

\bibitem{wei2023temporal}
W.~Wei, M.~Zhang, H.~Qu, A.~Belatreche, J.~Zhang, and H.~Chen, ``Temporal-coded spiking neural networks with dynamic firing threshold: Learning with event-driven backpropagation,'' in \emph{Proceedings of the IEEE/CVF International Conference on Computer Vision}, 2023, pp. 10\,552--10\,562.

\bibitem{fang2021incorporating}
W.~Fang, Z.~Yu, Y.~Chen, T.~Masquelier, T.~Huang, and Y.~Tian, ``Incorporating learnable membrane time constant to enhance learning of spiking neural networks,'' in \emph{Proceedings of the IEEE/CVF international conference on computer vision}, 2021, pp. 2661--2671.

\bibitem{wang2022ltmd}
S.~Wang, T.~H. Cheng, and M.-H. Lim, ``Ltmd: learning improvement of spiking neural networks with learnable thresholding neurons and moderate dropout,'' \emph{Advances in Neural Information Processing Systems}, vol.~35, pp. 28\,350--28\,362, 2022.

\bibitem{sohl2015deep}
J.~Sohl-Dickstein, E.~Weiss, N.~Maheswaranathan, and S.~Ganguli, ``Deep unsupervised learning using nonequilibrium thermodynamics,'' in \emph{International conference on machine learning}.\hskip 1em plus 0.5em minus 0.4em\relax PMLR, 2015, pp. 2256--2265.

\bibitem{ho2020denoising}
J.~Ho, A.~Jain, and P.~Abbeel, ``Denoising diffusion probabilistic models,'' \emph{Advances in neural information processing systems}, vol.~33, pp. 6840--6851, 2020.

\bibitem{dhariwal2021diffusion}
P.~Dhariwal and A.~Nichol, ``Diffusion models beat gans on image synthesis,'' \emph{Advances in neural information processing systems}, vol.~34, pp. 8780--8794, 2021.

\bibitem{rombach2022high}
R.~Rombach, A.~Blattmann, D.~Lorenz, P.~Esser, and B.~Ommer, ``High-resolution image synthesis with latent diffusion models,'' in \emph{Proceedings of the IEEE/CVF conference on computer vision and pattern recognition}, 2022, pp. 10\,684--10\,695.

\bibitem{li2023diffusion}
X.~Li, Y.~Ren, X.~Jin, C.~Lan, X.~Wang, W.~Zeng, X.~Wang, and Z.~Chen, ``Diffusion models for image restoration and enhancement--a comprehensive survey,'' \emph{arXiv preprint arXiv:2308.09388}, 2023.

\bibitem{shang2024resdiff}
S.~Shang, Z.~Shan, G.~Liu, L.~Wang, X.~Wang, Z.~Zhang, and J.~Zhang, ``Resdiff: Combining cnn and diffusion model for image super-resolution,'' in \emph{Proceedings of the AAAI Conference on Artificial Intelligence}, vol.~38, no.~8, 2024, pp. 8975--8983.

\bibitem{wu2023super}
Z.~Wu, X.~Chen, S.~Xie, J.~Shen, and Y.~Zeng, ``Super-resolution of brain mri images based on denoising diffusion probabilistic model,'' \emph{Biomedical Signal Processing and Control}, vol.~85, p. 104901, 2023.

\bibitem{gu2022vector}
S.~Gu, D.~Chen, J.~Bao, F.~Wen, B.~Zhang, D.~Chen, L.~Yuan, and B.~Guo, ``Vector quantized diffusion model for text-to-image synthesis,'' in \emph{Proceedings of the IEEE/CVF conference on computer vision and pattern recognition}, 2022, pp. 10\,696--10\,706.

\bibitem{kim2022diffusionclip}
G.~Kim, T.~Kwon, and J.~C. Ye, ``Diffusionclip: Text-guided diffusion models for robust image manipulation,'' in \emph{Proceedings of the IEEE/CVF conference on computer vision and pattern recognition}, 2022, pp. 2426--2435.

\bibitem{nichol2021glide}
A.~Nichol, P.~Dhariwal, A.~Ramesh, P.~Shyam, P.~Mishkin, B.~McGrew, I.~Sutskever, and M.~Chen, ``Glide: Towards photorealistic image generation and editing with text-guided diffusion models,'' \emph{arXiv preprint arXiv:2112.10741}, 2021.

\bibitem{okhotin2024star}
A.~Okhotin, D.~Molchanov, A.~Vladimir, G.~Bartosh, V.~Ohanesian, A.~Alanov, and D.~P. Vetrov, ``Star-shaped denoising diffusion probabilistic models,'' \emph{Advances in Neural Information Processing Systems}, vol.~36, 2024.

\bibitem{wu2024medsegdiff}
J.~Wu, R.~Fu, H.~Fang, Y.~Zhang, Y.~Yang, H.~Xiong, H.~Liu, and Y.~Xu, ``Medsegdiff: Medical image segmentation with diffusion probabilistic model,'' in \emph{Medical Imaging with Deep Learning}.\hskip 1em plus 0.5em minus 0.4em\relax PMLR, 2024, pp. 1623--1639.

\bibitem{mandlekar2021matters}
A.~Mandlekar, D.~Xu, J.~Wong, S.~Nasiriany, C.~Wang, R.~Kulkarni, L.~Fei-Fei, S.~Savarese, Y.~Zhu, and R.~Mart{\'\i}n-Mart{\'\i}n, ``What matters in learning from offline human demonstrations for robot manipulation,'' \emph{arXiv preprint arXiv:2108.03298}, 2021.

\bibitem{florence2022implicit}
P.~Florence, C.~Lynch, A.~Zeng, O.~A. Ramirez, A.~Wahid, L.~Downs, A.~Wong, J.~Lee, I.~Mordatch, and J.~Tompson, ``Implicit behavioral cloning,'' in \emph{Conference on Robot Learning}.\hskip 1em plus 0.5em minus 0.4em\relax PMLR, 2022, pp. 158--168.

\bibitem{shafiullah2022behavior}
N.~M. Shafiullah, Z.~Cui, A.~A. Altanzaya, and L.~Pinto, ``Behavior transformers: Cloning $ k $ modes with one stone,'' \emph{Advances in neural information processing systems}, vol.~35, pp. 22\,955--22\,968, 2022.

\bibitem{yao2023attention}
M.~Yao, G.~Zhao, H.~Zhang, Y.~Hu, L.~Deng, Y.~Tian, B.~Xu, and G.~Li, ``Attention spiking neural networks,'' \emph{IEEE transactions on pattern analysis and machine intelligence}, vol.~45, no.~8, pp. 9393--9410, 2023.

\end{thebibliography}
\end{document}